\documentclass{article}
\usepackage{spconf,amsmath,graphicx}
\usepackage[utf8]{inputenc}
\usepackage[table,xcdraw]{xcolor}
\usepackage{graphics,hyperref}
\usepackage[T1]{fontenc}
\usepackage{subcaption,float,amssymb,amsfonts,caption,booktabs,algorithmic,textcomp}
\title{PROBING STATISTICAL REPRESENTATIONS FOR END-TO-END ASR}
\name{Anna Ollerenshaw, Md Asif Jalal, Thomas Hain}
\address{Speech and Hearing Group, The University of Sheffield, Sheffield, UK\thanks{This work was partly supported by Voicebase Inc. at the Voicebase Research Center}}
\begin{document}
\ninept
\maketitle
\begin{abstract}
End-to-End automatic speech recognition (ASR) models aim to learn a generalised speech representation to perform recognition. In this domain there is little research to analyse internal representation dependencies and their relationship to modelling approaches. 
This paper investigates cross-domain language model dependencies within transformer architectures using SVCCA and uses these insights to exploit modelling approaches. It was found that specific neural representations within the transformer layers exhibit correlated behaviour which impacts recognition performance.

Altogether, this work provides analysis of the modelling approaches affecting contextual dependencies and ASR performance, and can be used to create or adapt better performing End-to-End ASR models and also for downstream tasks.
\end{abstract}
\section{Introduction}
\label{sec:intro}

The typical approach to develop a framework for ASR has been to use deep neural networks to recognise and align acoustic features to graphemes or phonemes; replacing the requirement for distinctly-optimised modules, such as acoustic, pronunciation or language models (LMs). Using End-to-End modelling approaches reduces the need for expert domain knowledge as it aims to jointly optimise the training regime while adapting to diverse speech environments. These factors have led to End-to-End ASR models becoming a popular choice for on-device deployment. 

Current research in End-to-End ASR modelling is dominated by three approaches: recurrent-transducers \cite{graves2012sequence}, Connectionist Temporal Classification models \cite{graves2006connectionist} or attention-based encoder-decoder architectures \cite{vaswani2017attention,sutskever2014sequence}. End-to-End ASR frameworks are typically dependant on the amount of data resources and are commonly fine-tuned to the corpora in order to improve the recognition performance, which has directed techniques to improve the representation capacity of the modelling approaches \cite{kim2014robust,park2019specaugment}. The ability to use larger amounts of training data significantly improves the recognition performance of End-to-End models \cite{luscher2019rwth}. However, previous work \cite{arpit2017closer} suggested that the relationship between memorisation and generalisation within these networks remains elusive, being referred to as a ``\textit{black-box}". It can be hypothesised that richer neural representations are not analogous to increasing neural depth or model size \cite{montufar2014number}. By learning representations of speech that are robust, the general recognition performance of the model should improve proportionately without the requirement for increasing model size or training data.

As the development of End-to-End ASR models continues, there is also a corresponding demand to be able to interpret the internal latent representations and the explainability of modelling approaches, such as the potential relationship between training data and the integration of LMs. Despite numerous variations of modelling approaches, there has also been little exploration of the internal model representations and their relationship to model recognition performance across different domains. Layer-wise analysis of models has been used to interpret modelling approaches and relationships between representations in multiple domains \cite{kornblith2019similarity,morcos2018insights,ollerenshaw2021insights}. SVCCA analysis techniques have been used to highlight neural representations with respect to their ability to generalise, by observing the relationship between the correlation coefficients of neural layers during training \cite{ollerenshaw2022insights}. 

As transformer modelling approaches achieve state-of-the-art results for End-to-End ASR, this work aims to understand how representations in transformer models adapt to out-of-domain LMs by analysing how the representations evolve across layers. The relationship between ASR performance and the neural representations is explored. Using unmatched sub-word LMs, it is possible to observe the dependencies of the representations within model layers and observe the impact of the variations. Section \ref{sec:tuning} shows that observing the representation dependencies is important to develop intuitive modelling approaches and improve recognition performance. 

The analysis methodology is first defined in Section \ref{sec:corr}, then the developed framework for analysis experiments is described in Section \ref{sec:exp_e2e}. The implementation of cross-domain LMs in the End-to-End ASR models is defined in Section \ref{sec:lang} and the analysis experiments are shown in Section \ref{sec:lang_exp}. Experiments analysing the adaptation of transformer model parameters are conducted in Section \ref{sec:tuning} with the results discussed in further detail in Section \ref{sec::discussion}. It was found that deeper transformer model layers contain learned representation dependencies for cross-domain LMs and recognition performance can be improved by \textit{tuning} the parameters to the hierarchical dependencies.

%The main findings are:
%Previous work from \cite{ollerenshaw2021insights} used the correlation index SVCCA, described in Section \ref{sec:corr}, to provide insights on the neural representations within End-to-End ASR models of varying dimensionality. 

% Add part about freezing layers
\vspace{-2mm}
\section{Correlation Analysis Methodology for End-to-End ASR Models}\label{sec:corr}

To map the posterior distribution of $p(Y|X)$ jointly, End-to-End ASR modelling approaches aim to approximate function of acoustic input $\mathcal{X}=\{x_1,...,x_T\}$ of length $\mathcal{T}$ over output labels $\mathcal{Y}=\{y_1,...,y_N\}$. The dependencies of the End-to-End model parameters, are not as clearly distinguished as traditional modular approaches to acoustic and language modelling, as they are not separately defined. These \textit{``black-box"} models are complex and it can be difficult to understand the relationship between the internal parameter dependencies and the resulting performance across datasets. By using SVCCA analysis from \cite{raghu2017svcca}, it is possible to observe a relationship between two sets of observations by the maximum correlations between the linear projections. This correlation analysis method is used to measure the linear relationship between vectors representing neural layers. For ASR models, the  activation output of a layer $z^l_i = \{z_i^l(x_1),...,(z_i^l(x_N)\}$ are extracted for dataset $\mathcal{X}$ and neuron $i$ in layer $l$. To observe the similarity between or within neural layers and to compare the relationship of the internal representation dependencies upon different variables, the use of different LMs and model parameters are varied and the similarities between layer representations are compared to derive insights regarding their relationship.

To measure correlation for $\mathcal{N}$ datapoints, pairs of vectors are sampled from views of  layers $l_1$ and $l_2$, which are projected, where $l_1=\{z_1^{l_1},...,z_{N_1}^{l_1}\}$ and $l_2=\{z_2^{l_2},...,z_{N_2}^{l_2}\}$. 
%The objective of canonical correlation analysis (CCA) is to maximise the correlation between two matrices when the original matrices are projected onto bases $v, \; s$:
%to attempt to define the maximum correlation direction between the linear projections
%\begin{equation}
%    \frac{v^T\sum_{XY}s}{\sqrt{v^T\sum_{XX}v}\sqrt{s^T\sum_{YY}s}}
%\end{equation}
%where $\sum_{XX}, \; \sum_{XY}, \; \sum_{YY}$ are the cross-covariance and covariance of the matrices. 
The projected views of $l_1$ and $l_2$ are pruned by the application of singular vector decomposition (SVD) to retain 99\% of the representative dimensions and to reduce the impact of potential noise. The application of SVD forms subspaces $l_1' \subset l_1$ and $l_2' \subset l_2$ and then CCA can then be applied to find vectors $v$ and $s$ that maximise correlation $\rho$ between the projections of $l_1'$ and $l_2'$:

\begin{equation}
    \rho = \frac{\langle v^T l_1',s^Tl_2'\rangle}{||v^T l_1'||\; ||s^T l_2'||}
\end{equation}

where $v, \; s$ are transforms that aim to maximise the correlation of the vectors. $\rho$ increases where neural representations have encoded more similar information. Further details regarding SVCCA analysis can be found in \cite{raghu2017svcca}.

\vspace{-2mm}
\section{Experimental Setup}\label{sec:exps}
\subsection{Transformers for ASR}\label{sec:tran_arch}
Transformers, initially published in \cite{vaswani2017attention}, are a widely chosen encoder-decoder architecture for speech recognition frameworks due their ability to parallelise the training regime. This enables use of larger amounts of training data. 
The encoder structure is comprised of stacked self-attention and point-wise, fully connected layers in blocks, with each block managed by a multi-head self-attention layer and feed forward layer. The attention module is a major component of the architecture as it enables pairwise position measurement for windows of the input speech. 
%The attention mechanism can be described by:
%\begin{equation}
%    Attention(Q,V,K)=softmax(\frac{QK^T}{\sqrt{d_k}V})
%\end{equation}
%where $\mathcal{Q}$ contains the query, $\mathcal{V}$ contains the values and $\mathcal{K}$ contains the keys for the input sequence. 
After the attention vector has passed through the feed forward layer, the output is passed onto another self-attention layer within the decoder to compute the embedding vectors. A final attention mechanism attends over the embedding vectors, in order to compute the relationship between the input and output sequences, before being passed to the final linear unit. A softmax function is then calculated over the target output.

A CNN front-end is incorporated with the transformer layers for feature extraction. The final convolutional layer was then projected to 12 stacked transformer encoder blocks with embedding dimensions of $512 \times 2048$ and 6 decoder layers with positional embeddings.

\vspace{-2mm}
\subsection{Language Modelling in End-to-End ASR}\label{sec:lang}
The integration of LMs within End-to-End architectures is relatively common-place as they can be used to supervise the training optimisation and also for decoding to aid recognition performance \cite{watanabe2018espnet,wang2019espresso}. However, it is unclear how the internal dependencies of the End-to-End models handle latent LM representations and whether there are similar learned representation spaces that are robust across different domains. By training models with cross-domain LMs, it is possible to observe these dependencies by comparing models with SVCCA analysis. 

For the analysis experiments, a sub-word LM is integrated by shallow-fusion decoding \cite{gulcehre2015using} and label smoothing \cite{szegedy2016rethinking} techniques. The sub-word LM is a 3 layer LSTM model. Shallow-fusion decoding computes the weighted sum of a pair of posterior distributions over sub-words; using one from the ASR model and one from the sub-word LM. The sub-word LM is an LSTM-based LM trained with restricted computational complexity, by only keeping the most frequent sub-words and splitting the rest into characters, to enable conversion with low information loss. Label smoothing computes the cross entropy loss during the model's training regime with a weighted mix of distributions from a unigram LM and one-hot targets from the dataset.

\vspace{-2mm}
\subsection{Correlation Analysis Framework}\label{sec:exp_e2e}

%It is unclear how the internal structures of End-to-End ASR models and their learned representation space are dependent upon techniques to integrate LMs and their relationship to performance. 
The framework developed in \cite{ollerenshaw2021insights} was utilised to investigate the relationships between internal dependencies and save the models during training. For all the experiments, the transformer models and LMs were trained using the ESPRESSO framework \cite{wang2019espresso}. The analysis was conducted for all models by extracting the activation outputs of each neural layer of the encoder for each training epoch. Each model was saved throughout all epochs and then a controlled input of 100 frames of unseen speech data was fed through the layers, whilst simultaneously extracting the activation outputs for each layer. 80-dimensional log Mel acoustic features with additional pitch features were extracted, from 25ms windows with a stride of 10ms, for all models.
%To ensure the functionality of the correlation index for layers of different lengths, the spatial dimensions of the neural layers were linearly interpolated, bringing the narrower layers of the models to the same dimensionality as wider layers.

\vspace{-2mm}
\subsubsection{Data}\label{sec:data}
For the experiments, three common US-English datasets from differing domains for ASR were chosen: Switchboard \cite{godfrey1992switchboard}, Librispeech \cite{panayotov2015librispeech} and WSJ \cite{paul1992design}. 
%The datasets are US-English to avoid the impact of cross-lingual variations upon recognition as this is out-of-scope for these experiments. 
The Switchboard dataset contains conversational telephone speech, Librispeech is a compilation of read audiobooks, and WSJ contains read news. 
%This variation of domains is hypothesised to lead to the generation of speaker-specific and domain-specific contextual embedding representation during training. 
The test sets for the Switchboard dataset, referred to as \textit{Swbd} and \textit{Callhome}, are derived from the LDC2002S09 set and contain 20 unreleased telephone conversations from Switchboard and 20 telephone unscripted conversations from Callhome. To ensure the transformer models were converged by training on comparative data to the LM, the model trained with Switchboard used up-sampled data (to 16kHz). The LMs trained with Librispeech were trained using the full 960 hour training set and the ASR models were tested on the \textit{test-clean} and \textit{test-other} sets. The training set for the WSJ LM was the \textit{si284} set, with the \textit{Dev93} set for validation and \textit{Eval92} for testing the ASR model performance. 

% --------------------------------------------------

% Below is an example of how to insert images. Delete the ``\vspace'' line,
% uncomment the preceding line ``\centerline...'' and replace ``imageX.ps''
% with a suitable PostScript file name.
% -------------------------------------------------------------------------
%begin{figure}[htb]

%\begin{minipage}[b]{1.0\linewidth}
%  \centering
%  \centerline{\includegraphics[width=8.5cm]{figures/residual dconv.png}}
%%  \vspace{2.0cm}
%  \centerline{(a) Result 1}\medskip
%\end{minipage}
%%
%\begin{minipage}[b]{.48\linewidth}
%  \centering
%  \centerline{\includegraphics[width=4.0cm]{figures/residual dconv.png}}
%%  \vspace{1.5cm}
%  \centerline{(b) Results 3}\medskip
%\end{minipage}
%\hfill
%\begin{minipage}[b]{0.48\linewidth}
%  \centering
%  \centerline{\includegraphics[width=4.0cm]{figures/residual dconv.png}}
%%  \vspace{1.5cm}
%  \centerline{(c) Result 4}\medskip
%\end{minipage}
%%
%\caption{Example of placing a figure with experimental results.}
%\label{fig:res}
%%
%\end{figure}

%\vfill
%\pagebreak
\vspace{-2.0mm}
\section{Cross-domain Language model Experimentation}
\label{sec:lang_exp}

\begin{figure}[ht]
    \centering
    \begin{subfigure}{\linewidth}
        \includegraphics[height=0.5\linewidth,width=\linewidth]{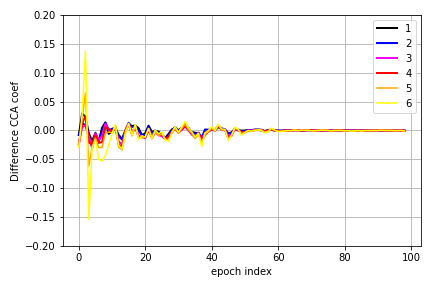}
    \end{subfigure}
    \begin{subfigure}{\linewidth}
        \includegraphics[height=0.5\linewidth,width=\linewidth]{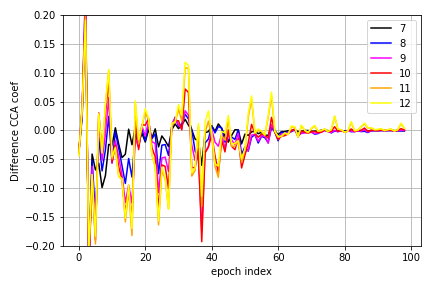}
    \end{subfigure}
    \caption{SVCCA correlation coefficients as performance converges within transformer layers 1 to 6 (\textbf{top}) and layers 6 to 12 (\textbf{bottom}), between a model trained with a Fisher-based LM and a model trained with a WSJ-based LM}
    \label{fig:lm_cca}
\end{figure}

Correlation analysis of the neural representations across the transformer model layers is used to measure and analyse the changes in correlation when cross-domain LMs are integrated. 
Figure \ref{fig:lm_cca} shows the SVCCA coefficients, as training converges, between the encoder layers of two transformer models. The models were trained with Switchboard data but one model uses an in-domain Fisher sub-word LM, and the other model uses an out-of-domain WSJ sub-word LM. These models are both trained with sub-word units using SentencePiece \cite{kudo2018subword} and integrated during the training process using the scheduled sampling method and decoded with shallow-fusion, as described in Section \ref{sec:lang}. The correlation analysis shows very little difference in coefficiency between layers 1 to 6 (top graph of Figure \ref{fig:lm_cca}), aside from in the initial epochs which could be attributed to the random initialisation of parameters. This suggests that the neural layers of both of these models are converging to similar representation spaces. However, between layers 7 to 12 (bottom graph of Figure \ref{fig:lm_cca}, the differences in coefficiency are much larger throughout training. This suggests that the representations learned in these deeper layers are more dependent upon the LM domain. 

\begin{figure}[ht]
    \centering
    \begin{subfigure}{\linewidth}
        \includegraphics[height=0.5\linewidth,width=\linewidth]{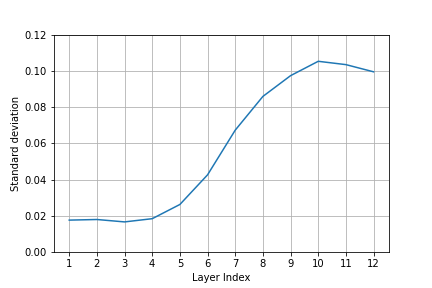}
    \end{subfigure}
    \begin{subfigure}{\linewidth}
        \includegraphics[height=0.5\linewidth,width=\linewidth]{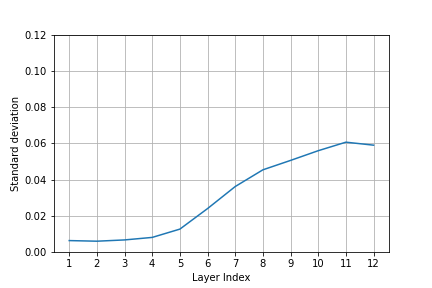}
    \end{subfigure}
    \caption{Standard deviation of correlations across transformer model layers with and without a LM (\textbf{top}) and with cross-domain LMs (\textbf{bottom})}
    \label{fig:stdev}
\end{figure}

The bottom graph in Figure \ref{fig:stdev} displays the standard deviations of the coefficiency between the models trained with cross-domain LMs. This aims to show the variation in coefficiency by layer more clearly, where the standard deviations in layers 10, 11 and 12 are highest. The top graph of Figure \ref{fig:stdev} shows the variance in coefficiency within the neural layers of a model trained without scheduled sampling or shallow-fusion decoding compared to the model trained with the Fisher sub-word LM. This suggests that a similar observation can be made for LM specific representations, whereby the variance is higher overall and the coefficiency of layers 8 to 12 deviates the most. The results in Figure \ref{fig:stdev} also imply that layers 1 to 4 have very little dependency on LM representations. These insights suggest that encoder layers 1 to 4 of the transformer model can be frozen when fine-tuning with LMs and the optimisation regime of End-to-End ASR models can be adapted to improve downstream tasks.


Regarding performance, the model that was trained with the Fisher LM reached 9.5\% word error rate (WER) on the Switchboard test set and 19.1\% on the Callhome test set, while the model that was trained with the WSJ LM was 10.7\% and 21.1\% respectively. The differences in recognition performance are attributed to the domains of the LMs and the test sets used for evaluation.

%%%%%%%%%%%%%%%%%%%%%%
%Graph of WSj, fisher LM vs no LM integration showing layers 1 to 5 converge without correlation differences after epoch 10. Suggests language model context learned in higher transformer layers.
\vspace{-2mm}
\section{Modelling Structure Analysis}
\label{sec:tuning}

To optimise the parameters for state-of-the-art End-to-End ASR models, many iterations are trained with slight parameter modifications. Optimisation of model parameters to specific datasets to achieve the best recognition performance possible \cite{wang2019espresso, watanabe2018espnet} is referred to here as \textit{tuned}. For example, the dimensionality, number of layers and also the hyperparameters have been observed to impact the recognition performance. As shown in Table \ref{tab:arch}, using a transformer model with the same parameters and composition for several datasets does not achieve the lowest WER across all of the datasets. These \textit{tuned} models are typically reached by extensive hyperparameter optimisation techniques, which are computationally expensive and considerably time consuming without providing observational evidence regarding the dependencies of certain parameters upon the recognition performance.

Using cross-corpora correlation analysis while varying the parameters, it is possible to interpret these dependencies in a more meaningful way and provide some observational evidence to reduce the future need for extensive hyperparameter optimisation when developing new models or fine-tuning trained models. By understanding the representation dependencies, this can potentially reduce the computational resources required to improve speech recognition model architectures. Table \ref{tab:arch} shows the results of 3 transformer models with variations in model parameters that are used in state-of-the-art End-to-End ASR frameworks. All models are the same transformer-based encoder-decoder architecture with the following variations: 
\begin{itemize}
    \item \textbf{Model 1} has an embedding dimension of 512, a feed forward embedding dimension of 2048, 4 attention heads, and an attention dropout of 0.25.
    \item \textbf{Model 2} has an embedding dimension of 256, a feed forward embedding dimension of 1024, 4 attention heads, and attention dropout of 0.25.
    \item \textbf{Model 3} has an embedding dimension of 512, a feed forward embedding dimension of 2048, 8 attention heads, and an attention dropout of 0.1. 
\end{itemize}
%These variations in model structures are the result of tuning to various datasets: Model 1 is \textit{tuned} to Switchboard and achieves a WER of 11.0\% and 20.7\% on the EVAL'00 test set; Model 2 is tuned to WSJ and achieves 4.13\% WER on the EVAL92 test set and 6.3\% on the DEV93 test set; Model 3 is tuned to Librispeech and achieves 1.9\% on Test-clean and 3.9\% on Test-other. 

To observe the relationship between the learned representations of the adapted models and attribute these adaptations to improved recognition performance with specific data, the model performance was assessed across all test sets, as shown in Table \ref{tab:arch}. For the Switchboard and Callhome test sets, the recognition performance of model 1 is the best, while model 2 reaches slightly worse performance on the Callhome set and model 3 has the highest WER for both test sets.

\begin{table}[ht!]
  \caption{Transformer model WER on EVAL'00, WSJ and Librispeech test sets with \textit{tuned} parameters}
  \label{tab:arch}
  \centering
  \resizebox{\columnwidth}{!}{%
  \begin{tabular}{c c c c c c c}
  \toprule
  \multicolumn{1}{c}{\textbf{Model}} & \multicolumn{1}{c}{\textbf{Swbd}} & \multicolumn{1}{c}{\textbf{Chm}} &
  \multicolumn{1}{c}{\textbf{Eval92}} &
  \multicolumn{1}{c}{\textbf{Dev93}} &
  \multicolumn{1}{c}{\textbf{Test-cln}} &
  \multicolumn{1}{c}{\textbf{Test-oth}}\\
    \midrule
    M1 & \textbf{9.5} & \textbf{19.1} & 4.59 & 7.54 & 3.5 & 8.51 \\
    M2 & 9.6 & 20 & \textbf{4.13} & \textbf{6.3} & 3.99 & 8.72 \\
    M3 & 10.4 & 21.6 & 4.52 & 7.43 & \textbf{1.9} & \textbf{3.9}  \\
    \bottomrule
  \end{tabular}%
  }
\end{table}

\begin{figure}[ht]
    \centering
    \begin{subfigure}{\linewidth}
        \includegraphics[height=0.5\linewidth,width=\linewidth]{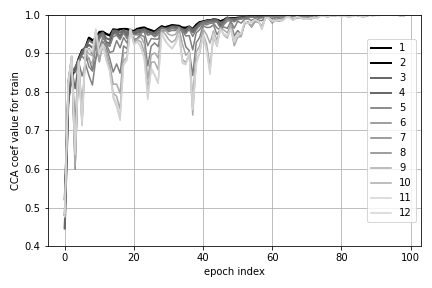}
    \end{subfigure}
    \begin{subfigure}{\linewidth}
        \includegraphics[height=0.5\linewidth,width=\linewidth]{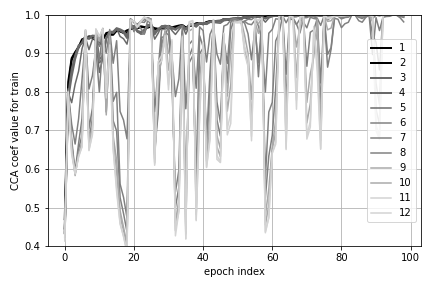}
    \end{subfigure}
    \begin{subfigure}{\linewidth}
        \includegraphics[height=0.5\linewidth,width=\linewidth]{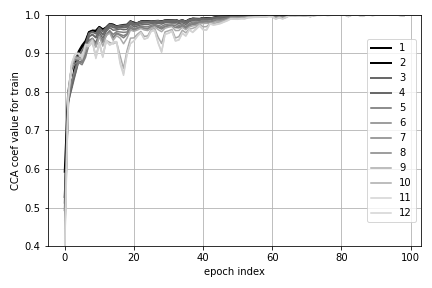}
    \end{subfigure}
    \caption{Transformer SVCCA coefficients as performance converges in model 1 (top), model 2 (middle) and model 3 (bottom) trained with Switchboard data}
    \label{fig:libri_arch}
\end{figure}

Figure \ref{fig:libri_arch} displays the SVCCA coefficients for each model trained with the Switchboard dataset. Model 2's mean coefficiency, across layerwise representations, are substantially less correlated than the other models. The standard deviations of the correlations within these layers also vary significantly higher than Model 1 or 3. Model 3's mean coefficiency across layerwise representations is fairly similar for all layers with very small standard deviation. It is observed that correlations within the layers of model 2 have lower coefficiency, and the recognition performance of this model is lower for the Switchboard test sets. Also, there are little hierarchical coefficiency patterns throughout the layers of model 3, and this model also has a slightly worse performance, which corroborates with results from \cite{ollerenshaw2021insights}. Model 1 has lower coefficiency within layers 8-12 and has the best recognition performance. 

%\begin{figure}[h]
%    \centering
%    \includegraphics[width=\linewidth,height=0.6\linewidth]{figures/mean_bar.png}
%    \caption{Average coefficiency correlations across architecture layers (bars) and standard deviations in %coefficiency correlation across architecture layers (lines)}
%    \label{fig:tuned_bar}
%\end{figure}

% --------------------------------------------------
\vspace{-2mm}
\section{Discussion}
\label{sec::discussion}

The findings in Section \ref{sec:lang} correlate with findings from \cite{shah2021all} where semantic and syntax level features of speech are predominantly dependent upon deeper layers of transformer-based models, while acoustic and fluency features are predominantly dependent on the shallower layers. In the case of End-to-End transformers for ASR tasks, the LM-dependent representations are shown to be primarily dependent within layers 7-12. The cross-domain LM-dependent representations are observed within layers 10-12. Further experiments training with the WSJ dataset with Fisher and WSJ sub-word LMs showed very similar observed behaviours across layer coefficiency. These observations can be used to check for possible biases in the modelling process that affect recognition performance, without the need for extensive training requirements, to improve joint optimisation. The analysis also aids in the interpretability of the impact of representation dependencies within End-to-End ASR models. 

The experiments in Section \ref{sec:tuning}, attempt to show these internal dependencies with regard to the model parameters within the same modelling approaches. As shown in Figure \ref{fig:libri_arch}, Model 2 used shallower embedding dimensions than model 1, which has caused the coefficiency of many of the layers to become highly uncorrelated. Model 3 is observed to have very highly correlated layers, however there are little distinct hierarchies in the neural representations when the attention heads are increased to 8 and the attention dropout is reduced. By adapting the parameters of transformer models, the layers with the most dependency for representing domain-specific information are altered. These changes in hierarchical representations have been observed to impact recognition performance, and further suggests a relationship between correlated hierarchical representations and the ability for the model to generalise, particularly for cross-domain speech recognition. Increasing the attention dropout is theorised to improve model robustness \cite{park2019specaugment}, where typical features of conversational speech are boundary uncertainties and hesitations. In the case of End-to-End conversational speech recognition, the results show that using substantial attention dropout in transformer models is important to produce correlated hierarchies in dependent layers but also utilise a model with sufficient embedding dimensionality that the representations within context-critical layers don't become too uncorrelated. 

% asif: this section is unclear 
% ao Increasing the dropout for a model training with conversational telephone speech is theorised to improve model robustness to typical features of this type of speech, such as boundary uncertainties and hesitations.
%%%%%%%%%%%%%%%%%%%%%%%%%%%%%%%%%%%%%%%%%%%%%%%%%%

\vspace{-2mm}
\section{Conclusion}\label{sec:conclusion}
Using SVCCA as a correlation index has highlighted several aspects of the relationships between the neural representations, transformer-based modelling parameters and the impact these have upon recognition performance. Interpretative analysis is important to develop future modelling approaches for meaningful improvement strategies. Expanding the scope of the investigation into the attributes and potential learned features that could be classified within the layers would provide a deeper understanding of the properties of these dependencies and how these could be further exploited. The insights into the dependencies of the neural layers can be used for the development of models for few-shot learning and downstream tasks for End-to-End ASR. 

% -------------------------------------------------------------------------

%\section{Acknowledgements}
%This work was partly supported by Voicebase Inc. at the Voicebase Research Center.
\newpage
\bibliographystyle{IEEEbib}
\bibliography{main}

\end{document}